%% file: main.tex
\documentclass[conference]{IEEEtran}
\IEEEoverridecommandlockouts
\usepackage{cite}
\usepackage{amsmath,amssymb,amsfonts}
\usepackage{algorithmic}
\usepackage{graphicx}
\usepackage{textcomp}
\usepackage{xcolor}
\usepackage{hyperref}
\usepackage{booktabs}
\usepackage{algorithm}
\usepackage{subcaption}
\usepackage{tikz}
\usetikzlibrary{shapes.geometric, arrows.meta, positioning, fit, backgrounds, calc}
\def\BibTeX{{\rm B\kern-.05em{\sc i\kern-.025em b}\kern-.08em
    T\kern-.1667em\lower.7ex\hbox{E}\kern-.125emX}}
\begin{document}

\title{An Agentic LLM Framework for\\Adverse Media Screening in AML Compliance}

\author{\IEEEauthorblockN{Pavel Chernakov, Sasan Jafarnejad, Rapha\"el Frank}
\IEEEauthorblockA{\textit{Interdisciplinary Centre for Security, Reliability and Trust}\\
University of Luxembourg\\
\{pavel.chernakov, sasan.jafarnejad, raphael.frank\}@uni.lu}
}

\maketitle
\input{sections/00_abstract}
\input{sections/01_introduction}
\input{sections/02_related_work}

\input{sections/03_system_architecture}

\input{sections/04_methodology}

\input{sections/05_results}
\input{sections/06_discussion}
\input{sections/07_conclusion}

\section*{Acknowledgment}
This research was funded in whole, or in part, by the Luxembourg National Research Fund (FNR), grant reference NCER22/IS/16570468/NCER-FT. For the purpose of open access, and in fulfilment of the obligations arising from the grant agreement, the author has applied a Creative Commons Attribution 4.0 International (CC BY 4.0) license to any Author Accepted Manuscript version arising from this submission.

\end{document}

%% file: sections/00_abstract.tex
\begin{abstract}
Adverse media screening is a critical component of anti-money laundering (AML) and know-your-customer (KYC) compliance processes in financial institutions. Traditional approaches rely on keyword-based searches that generate high false-positive rates or require extensive manual review. We present an agentic system that leverages Large Language Models (LLMs) with Retrieval-Augmented Generation (RAG) to automate adverse media screening. Our system implements a multi-step approach where an LLM agent searches the web, retrieves and processes relevant documents, and computes an Adverse Media Index (AMI) score for each subject. We evaluate our approach using multiple LLM backends on a dataset comprising Politically Exposed Persons (PEPs), persons from regulatory watchlists, and sanctioned persons from OpenSanctions and clean names from academic sources, demonstrating the system's ability to distinguish between high-risk and low-risk individuals.
\end{abstract}

\begin{IEEEkeywords}
adverse media, anti-money laundering, LLM agents, RAG, compliance automation, AMLR
\end{IEEEkeywords}

%% file: sections/01_introduction.tex
\section{Introduction}

Adverse media screening, also known as negative news screening, refers to the process of identifying publicly available information that may indicate a customer's involvement in financial crime, corruption, fraud, or other illicit activities. This screening is a fundamental component of customer due diligence (CDD) and enhanced due diligence (EDD) processes required by anti-money laundering (AML) regulations worldwide~\cite{fatf2012}. In our prior work, we introduced a risk-based framework for Know Your Customer (KYC) processes that incorporates multiple risk dimensions~\cite{jafarnejad2024riskkyc}; the present work extends this framework by focusing specifically on the adverse media screening component.

The importance of adverse media screening has grown significantly with the evolution of European AML regulations. The 4th Anti-Money Laundering Directive (4AMLD) introduced requirements for screening against open-source media, specifically ``reports in reputable newspapers.'' The 5th AMLD expanded the scope of business sectors required to perform such searches and introduced requirements for automation in adverse media screening processes. The 6th AMLD further extended the framework by adding new money laundering predicate offences, including cybercrime and environmental crime, necessitating adjustments to adverse media screening to cover these new categories~\cite{ripjar2023}.

Notably, the European regulatory landscape is undergoing a significant transformation with the introduction of the Anti-Money Laundering Regulation (AMLR), which will supersede many provisions of the existing AMLD framework. The AMLR, expected to be directly applicable across all EU member states without requiring national transposition, aims to create a more harmonised and stringent AML framework. This regulatory evolution underscores the need for flexible, adaptable adverse media screening systems that can accommodate evolving compliance requirements~\cite{eu2024amlr}.

Traditional approaches to adverse media screening present significant challenges. Adverse media is generally understood as information available in the public domain which financial institutions would consider relevant to the management of financial crime risk. However, operationalizing this definition at scale remains problematic. Current approaches can be categorized into three main types~\cite{complyadvantage2023}:

\begin{enumerate}
    \item \textbf{Keyword-based approaches}: These systems search the web using predefined keywords associated with financial crime. While cost-efficient, they generate high false-positive rates due to the signal-to-noise problem, where legitimate mentions are conflated with genuinely adverse information.
    
    \item \textbf{Hybrid approaches}: Combining keyword searches with human expert review, these systems achieve better precision but at significant cost and with limited scalability. Human reviewers must manually filter results and assess context, creating bottlenecks in high-volume screening operations.
    
    \item \textbf{Machine learning approaches}: These systems employ traditional natural language processing (NLP) and machine learning techniques to classify and filter results. While more scalable than hybrid approaches, they require substantial training data and struggle with contextual understanding and entity disambiguation.
\end{enumerate}

We propose a fourth approach that leverages LLMs as autonomous agents capable of sophisticated contextual reasoning. Our system, which we call the AMI Agent, implements a complete adverse media screening pipeline that operates as follows:

\begin{enumerate}
    \item Performs web searches for a given identity using configurable search APIs
    \item Retrieves and parses relevant web documents, extracting textual content
    \item Applies RAG to identify relevant document segments
    \item Executes a structured playbook of assessment questions using the LLM
    \item Computes an aggregate Adverse Media Index score with natural language justifications
\end{enumerate}

The key contributions of this work are:
\begin{itemize}
    \item An agentic architecture for automated adverse media screening using LLMs with configurable assessment playbooks
    \item A multi-dimensional scoring approach that separately evaluates identity matching confidence, sentiment negativity, and activity risk level
    \item An open-source implementation supporting multiple LLM backends including local models (Ollama, vLLM) and API-based services (OpenAI, OpenRouter)
    \item Comparative evaluation across multiple LLM backends on four distinct populations.
\end{itemize}

The remainder of this paper is organized as follows: Section~\ref{sec:related} reviews related work in adverse media screening and LLM agents. Section~\ref{sec:architecture} describes our system architecture in detail. Section~\ref{sec:methodology} presents our experimental methodology. Section~\ref{sec:results} reports experimental results. Section~\ref{sec:discussion} discusses findings and limitations. Section~\ref{sec:conclusion} concludes the paper.

%% file: sections/02_related_work.tex
\section{Related Work}
\label{sec:related}

\subsection{Regulatory Framework for Adverse Media Screening}

The regulatory requirements for adverse media screening have evolved significantly over the past decade. The Financial Action Task Force (FATF) recommendations establish the international standards for AML and Combating the Financing of Terrorism (CFT) compliance, requiring financial institutions to conduct customer due diligence that includes understanding the nature and purpose of business relationships~\cite{fatf2012}.

At the European level, the progression from 4AMLD through 6AMLD has incrementally strengthened requirements. The 4AMLD established the baseline requirement for adverse media checks, while 5AMLD explicitly required automation capabilities and expanded the scope to include cryptocurrency exchanges and wallet providers. The 6AMLD harmonized the definition of money laundering predicate offences across member states and introduced environmental crime and cybercrime as new categories~\cite{ripjar2023}.

The forthcoming AMLR represents a paradigm shift from directive-based to regulation-based governance. Unlike directives, which require transposition into national law, the AMLR will be directly applicable, eliminating inconsistencies in implementation across member states. The regulation is expected to introduce more prescriptive requirements for customer screening, including adverse media screening, and will be enforced by the new Anti-Money Laundering Authority (AMLA)~\cite{eu2024amlr}.

Regional guidance has also been provided by supervisory authorities. The Hong Kong Monetary Authority (HKMA) has issued guidance on transaction monitoring and customer screening~\cite{hkma2018}, while the Monetary Authority of Singapore (MAS) has published detailed guidance on effective AML/CFT controls~\cite{mas2018}. These guidelines, while not specifically focused on adverse media, establish expectations for risk-based approaches to customer screening.

\subsection{Adverse Media Screening Systems}

Prior work on automated adverse media screening has explored various technical approaches. The Adverse Media Screening Portal~\cite{amsp2023} demonstrated the feasibility of web-based screening interfaces but relied primarily on keyword matching with limited contextual analysis.

Commercial solutions from vendors such as Refinitiv World-Check\footnote{\url{https://www.lseg.com/en/risk-intelligence/screening-solutions/world-check-kyc-screening}}, Dow Jones Risk \& Compliance\footnote{\url{https://www.dowjones.com/professional/risk/}}, and LexisNexis\footnote{\url{https://risk.lexisnexis.com/}} provide comprehensive screening databases but operate as ``black box'' systems with limited transparency into scoring methodologies. These systems typically combine curated databases of known adverse entities with real-time media screening, but their proprietary nature limits academic evaluation and customization~\cite{complyadvantage2023}.

The signal-to-noise problem remains a central challenge in adverse media screening. Names are inherently ambiguous, common names may match thousands of individuals, while variations in spelling, transliteration, and aliases complicate entity resolution. Traditional systems address this through fuzzy matching algorithms and manual review workflows, but these approaches scale poorly with increasing screening volumes.

\subsection{Large Language Models and Agents}

The emergence of LLMs has opened new possibilities for natural language understanding tasks. Models such as GPT-4, Claude, and open-source alternatives like Llama and Mistral have demonstrated remarkable capabilities in contextual reasoning, summarization, and question answering~\cite{openai2023gpt4, touvron2023llama}.

The concept of LLM agents---systems that use LLMs as reasoning engines to autonomously execute multi-step tasks---has gained significant attention. The ReAct framework demonstrated that interleaving reasoning traces with action execution improves task performance~\cite{yao2023react}. Chain-of-thought prompting showed that explicit reasoning steps enhance LLM performance on complex tasks~\cite{wei2022chain}.

RAG addresses the knowledge limitations of LLMs by combining generative models with external retrieval systems~\cite{lewis2020rag}. In RAG architectures, relevant documents are retrieved from a corpus and provided as context to the LLM, enabling more accurate and grounded responses. This approach is particularly relevant for adverse media screening, where the relevant information exists in external web documents rather than the model's training data.

The DSPy framework provides a programmatic approach to constructing LLM pipelines with typed input/output signatures~\cite{khattab2023dspy}. This enables structured extraction of information from LLM responses, which is essential for computing quantitative risk scores from natural language assessments.

\subsection{Named Entity Recognition and Disambiguation}

Entity disambiguation---determining whether a name mention refers to a specific individual---is a longstanding NLP challenge relevant to adverse media screening. Traditional approaches use feature-based classifiers with contextual signals such as co-occurring entities, temporal references, and topical indicators~\cite{bunescu2006using}.

Recent work has applied neural approaches to entity linking, leveraging contextual embeddings to improve disambiguation accuracy~\cite{wu2020scalable}. However, these approaches typically require training data with ground-truth entity labels, which is scarce in the adverse media domain. Our approach addresses this by using LLMs for zero-shot entity matching, leveraging their pre-trained world knowledge to assess whether document content matches a target identity.

%% file: sections/03_system_architecture.tex
\section{System Architecture}
\label{sec:architecture}

\begin{figure*}[htbp]
\centering
\resizebox{\textwidth}{!}{%
\begin{tikzpicture}[
    node distance=0.9cm and 1.4cm,
    input/.style={rectangle, rounded corners, draw=black!70, fill=blue!10, thick, minimum height=1.1cm, minimum width=2.2cm, align=center, font=\small},
    module/.style={rectangle, rounded corners, draw=black!70, fill=orange!12, thick, minimum height=1.2cm, minimum width=2.4cm, align=center, font=\small},
    submodule/.style={rectangle, rounded corners, draw=black!40, fill=orange!5, minimum height=0.65cm, minimum width=1.7cm, align=center, font=\scriptsize},
    output/.style={rectangle, rounded corners, draw=black!70, fill=green!12, thick, minimum height=1.1cm, minimum width=2.2cm, align=center, font=\small},
    external/.style={cylinder, shape border rotate=90, draw=black!40, fill=gray!8, aspect=0.3, minimum height=0.9cm, minimum width=1.5cm, align=center, font=\scriptsize},
    arrow/.style={-{Stealth[length=2.5mm]}, thick, draw=black!70},
    dataarrow/.style={-{Stealth[length=2mm]}, draw=blue!40, densely dashed},
    label/.style={font=\scriptsize, text=black!60},
]


\node[input] (identity) {\textbf{Identity}\\{\scriptsize Name, DOB, ...}};

\node[module, right=of identity] (search) {\textbf{Search}\\{\scriptsize\textbf{Engine}}};

\node[module, right=of search] (crawler) {\textbf{Web}\\{\scriptsize\textbf{Crawler}}};

\node[module, right=of crawler] (docproc) {\textbf{Document}\\{\scriptsize\textbf{Processor}}};

\node[module, right=of docproc] (agent) {\textbf{LLM Agent}};

\node[module, right=of agent] (verdict) {\textbf{Verdict}\\{\scriptsize\textbf{Generator}}};

\node[output, right=of verdict] (output) {\textbf{AMI Score}\\{\scriptsize + Justification}};

\node[external, above=0.7cm of search] (searchapi) {Search API};
\node[external, above=0.7cm of crawler] (web) {Web};
\node[external, above=0.7cm of agent] (llm) {LLM Backend};

\node[submodule, below=0.6cm of docproc, xshift=-1.0cm] (chunk) {Chunking};
\node[submodule, below=0.6cm of docproc, xshift=1.0cm] (embed) {Embedding};
\node[external, below=1.6cm of docproc, xshift=1.5cm] (vectordb) {Vector Store};

\node[submodule, below=0.6cm of agent, xshift=-1.0cm] (playbook) {Playbook};
\node[submodule, below=0.6cm of agent, xshift=1.0cm] (rag) {RAG};

\draw[arrow] (identity) -- (search);
\draw[arrow] (search) -- (crawler);
\draw[arrow] (crawler) -- (docproc);
\draw[arrow] (docproc) -- (agent);
\draw[arrow] (agent) -- (verdict);
\draw[arrow] (verdict) -- (output);

\node[label, above=0.2cm of search.west, anchor=south west] {URLs};
\node[label, above=0.2cm of crawler.west, anchor=south west] {Pages};
\node[label, above=0.2cm of docproc.west, anchor=south west] {Docs};
\node[label, above=0.2cm of agent.west, anchor=south west] {Chunks};
\node[label, above=0.2cm of verdict.west, anchor=south west] {Evidence};

\draw[dataarrow] (search) -- (searchapi);
\draw[dataarrow] (crawler) -- (web);
\draw[dataarrow] (agent) -- (llm);

\draw[dataarrow] (docproc) -- (chunk);
\draw[dataarrow] (docproc) -- (embed);
\draw[dataarrow] (embed) -- (vectordb);
\draw[dataarrow] (agent) -- (playbook);
\draw[dataarrow] (agent) -- (rag);

\draw[dataarrow] (vectordb) -- (rag);

\end{tikzpicture}
}
\caption{System architecture showing the agentic adverse media screening pipeline. The pipeline processes an identity through five stages: (1) \textbf{Search Engine} queries web APIs for relevant URLs, (2) \textbf{Web Crawler} retrieves page content, (3) \textbf{Document Processor} chunks text and creates embeddings stored in a vector store, (4) \textbf{LLM Agent} executes configurable playbooks using RAG-based question answering, and (5) \textbf{Verdict Generator} synthesizes evidence into a final AMI score with justification.}
\label{fig:architecture}
\end{figure*}
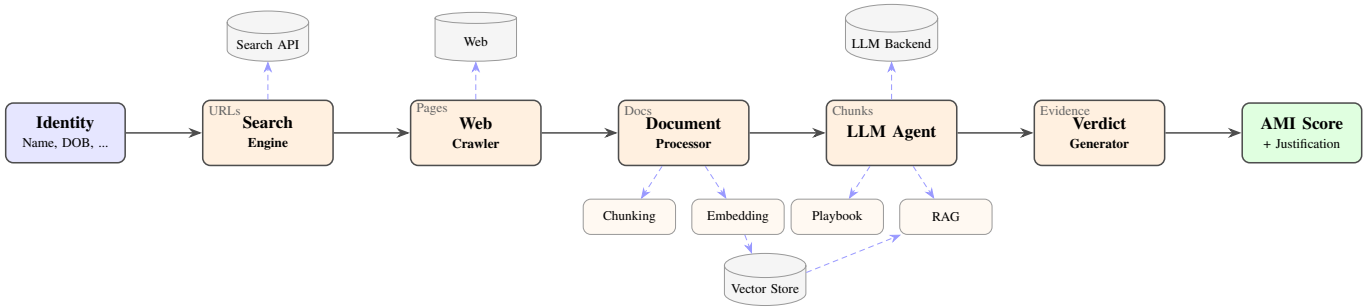

Our system implements an agentic pipeline for adverse media screening. The architecture, illustrated in Fig.~\ref{fig:architecture}, consists of four main components: Search Engine, Web Crawler, Document Processor, and LLM Agent.

\subsection{Search Engine Module}

The Search Engine module initiates the screening process by querying web search APIs for the target identity. We currently support the Google Custom Search API, which provides programmatic access to Google search results with configurable parameters.

Given an identity specification (minimally a name, optionally with additional metadata such as date of birth or known affiliations), the module constructs a search query and retrieves the top-$k$ results. Each result includes the page URL, title, and snippet. The module supports filtering to exclude irrelevant domains (e.g., social media profiles, generic directory sites, unsupported file formats) and can operate in snapshot mode for reproducible evaluations.

\subsection{Web Crawler Module}

The Web Crawler module retrieves and parses the HTML content of URLs identified by the search engine. We employ parsing using the LangChain library.

The crawler extracts the main textual content from each page, removing navigation elements, advertisements, and boilerplate text where possible. Extracted content is wrapped in LangChain Document objects that preserve source metadata (URL, title, retrieval timestamp) alongside the textual content.

Error handling addresses common issues including connection timeouts, HTTP errors, and malformed HTML. Pages that cannot be successfully parsed are logged and excluded from the set of resources used to compute the Adverse Media Index later on.

\subsection{Document Processor Module}

The Document Processor module prepares retrieved documents for analysis. This involves two main steps: chunking and embedding.

\textbf{Chunking}: Documents are split into overlapping chunks using a recursive character-based splitter. Chunk size is configurable, with defaults of 1000 tokens for API-based models and 500 tokens for local models (to accommodate smaller context windows or hardware limitations). An overlap of 10\% ensures that relevant information spanning chunk boundaries is captured.

\textbf{Embedding}: Each chunk is converted to a dense vector representation using an embedding model. We support multiple embedding backends:
\begin{itemize}
    \item Local models: BAAI/bge-large-en-v1.5, Qwen3-Embedding-0.6B, multilingual-e5-large-instruct
    \item API-based: embeddings via an OpenAI-compatible API
\end{itemize}

Embeddings are stored in a FAISS vector store, enabling efficient similarity search during the question-answering phase. The system maintains a cache of processed documents (keyed by content hash) to avoid redundant embedding computation.

\subsection{LLM Agent Module}

The LLM Agent module is the core reasoning component of our system. It executes structured assessment playbooks against the retrieved documents, using the LLM to answer questions and assign scores.

\subsubsection{Backend Support}

The agent supports multiple LLM backends through a unified interface:
\begin{itemize}
    \item \textbf{Ollama}: Self-hosted execution of open source models with compatibility for diverse hardware setups
    \item \textbf{vLLM}: Self-hosted high-throughput solution with support for a variety of open source models
    \item \textbf{OpenAI}: API access to GPT series models
    \item \textbf{OpenRouter\footnote{\url{https://openrouter.ai/}}}: Unified API access to multiple model providers
\end{itemize}

The backend options allow the developers to configure the system based on their needs in terms of aspects such as cost or latency. In case of hosting models in-house, the added benefit is data privacy, whereas the API options may provide easier and faster access to superior capabilities.

\subsubsection{DSPy Signatures}

We use the DSPy framework to define typed signatures for LLM interactions. Key signatures include:

\textbf{ConfidenceScore}: Given a context (document chunk) and question, returns a floating-point score (0-1) with a natural language justification.

\textbf{SuspicionVerdict}: Given accumulated evidence (questions, document-wise scores and justifications), returns a final verdict score with summary justification.

These typed signatures ensure structured outputs that can be reliably parsed and aggregated.

\subsubsection{Playbook Execution}

A playbook defines a sequence of assessment questions to evaluate a person based on the retrieved documents. The default playbook includes three dimensions:

\begin{enumerate}
    \item \textbf{Identity Matching (Question A)}: ``On a scale from 0 to 1, how confident are you that this document describes a person named [NAME]?''
    
    \item \textbf{Negativity Assessment (Question B)}: ``On a scale from 0 to 1, how negative is this description of [NAME]? Only consider explicitly stated facts.''
    
    \item \textbf{Risk Evaluation (Question C)}: ``On a scale from 0 to 1, how high-risk is the activity attributed to [NAME]? Only use explicitly stated information.''
\end{enumerate}

For each question and document pair, the agent retrieves the top-$k$ most relevant chunks from the vector store and combines them into a single query to the LLM to obtain a score and a justification. After the results are computed for each document, a question-level mean is produced given all the individual document scores.

The playbook structure is configurable via YAML, allowing customisation for different screening scenarios (e.g., adding questions for specific risk categories and adjusting score scales).

\subsection{Verdict Generator Module}

The Verdict Generator synthesizes playbook results into a final Adverse Media Index score. It receives the accumulated evidence---all questions, per-document scores, and justifications---and produces:

\begin{itemize}
    \item A final AMI score (0-1 scale)
    \item A natural language summary explaining the score
\end{itemize}

The verdict prompt asks: ``How high-risk is doing business with this person, based on the context?'' The LLM considers the consistency of evidence across documents, the severity of identified issues, and the confidence of identity matching.

\subsection{Pipeline Orchestration}

The complete pipeline is orchestrated by a runner class that coordinates the modules.




%% file: sections/04_methodology.tex
\section{Methodology}
\label{sec:methodology}

\subsection{Dataset Construction}

We evaluate our system on four populations of subjects designed to represent different points along the risk spectrum:

\subsubsection{Clean Names (Low-Risk)}

To evaluate false-positive rates, we require a population of individuals unlikely to have genuine adverse media. We source ``clean'' names from DBLP, a comprehensive database of computer science publications. Specifically, we query for top-publishing authors at major venues (e.g., NeurIPS), reasoning that prolific academic researchers are unlikely to be associated with financial crime.

This population provides negative examples---individuals for whom our system should produce low AMI scores. Any elevated scores for clean names indicate potential false positives requiring investigation.

\subsubsection{PEP Names (Low/Moderate-Risk)}

PEPs are individuals who hold or have held prominent public functions, making them higher-risk for potential involvement in corruption or money laundering. We source PEP names from the OpenSanctions dataset, an open-source database aggregating sanctions lists, PEP databases, and other watchlists from official sources worldwide.

The OpenSanctions PEP dataset includes names, aliases, dates of birth, nationalities, and positions held. We sample names with Latin-script representations to ensure compatibility with English-language web search. This population represents individuals for whom adverse media is likely to exist, providing positive examples for our screening system.

\subsubsection{Regulatory Watchlist Names (Moderate-Risk)}

Persons on Regulatory Watchlists (RW) are individuals that are subject to restrictions and/or regulatory actions. These may include investor warnings, penalties, fines, or bans from specific activities or industries. The dataset is obtained from the OpenSanctions Regulatory Watchlists collection, which aggregates data from 32 sources including enforcement actions from financial regulators such as the US SEC, FINRA, and various European authorities. Names on this list are more likely to be linked to illicit activities in comparison to PEPs.

\subsubsection{Sanctioned Names (High-Risk)}

Sanctioned Persons are taken from the list of Specially Designated Nationals (SDNs) maintained by the US Office of Foreign Assets Control (OFAC) and represent the cohort most likely (and oftentimes confirmed) to be involved in illicit activities beyond financial crime.

\subsection{Experimental Protocol}

For each name in our evaluation set, we execute the following protocol:

\begin{enumerate}
    \item Execute web search and retrieve top-10 results
    \item Parse and chunk retrieved documents
    \item Execute the default 3-question playbook
    \item Generate final verdict score
    \item Record all scores, justifications, and metadata
\end{enumerate}

To ensure reproducibility, we support snapshot-based evaluation where search results are cached and reused across experiments. This eliminates variability from search engine result changes over time.

\subsection{Model Configurations}

We evaluate three LLM configurations: Grok 4.1 Fast, Gemini 2.5 Flash, and GPT-4.1-Mini all via OpenRouter.

For embeddings, we primarily use Qwen3-Embedding-0.6B, a high-performing open-source embedding model suitable for retrieval tasks.

\subsection{Evaluation Metrics}

We assess system performance using the following metrics: \textit{Score Separation}, \textit{Distribution Analysis}, \textit{Justification Quality}, and \textit{Efficiency}. \textit{Score Separation} considers the difference between mean AMI scores for the four distinct populations. 
\textit{Distribution Analysis} is performed by visualizing score distributions to assess overlap between populations and identify optimal decision thresholds. \textit{Justification Quality} assesses whether justifications accurately reflect document content and logically support assigned scores. \textit{Efficiency} measures processing time and API cost per screening, relevant for production deployment considerations.





%% file: sections/05_results.tex
\section{Results}
\label{sec:results}

We evaluated our AMI Agent system using three state-of-the-art LLM backends: Grok 4.1 Fast, Gemini 2.5 Flash, and GPT-4.1-Mini. The evaluation was conducted across the four populations described in Section~\ref{sec:methodology}: Clean, PEP, RW, and SDN.

Table~\ref{tab:results} presents the mean AMI scores for each model across the four populations. All three models demonstrate clear separation between low-risk (Clean) and high-risk (SDN) populations, with SDN scores ranging from 0.73 to 0.863 compared to Clean scores of 0.015 to 0.029. The intermediate populations (PEP and RW) show appropriately graduated risk levels.

\begin{table}[htbp]
\caption{Mean AMI Scores by Model and Population}
\begin{center}
\begin{tabular}{lcccc}
\toprule
\textbf{Model} & \textbf{Clean} & \textbf{PEP}  & \textbf{RW}& \textbf{SDN} \\
\midrule
Grok 4.1 Fast & 0.015 & 0.073 & 0.179 & 0.863 \\
Gemini 2.5 Flash & 0.028 & 0.087 & 0.169 & 0.819 \\
GPT-4.1-Mini & 0.029 & 0.063 & 0.167 & 0.730 \\
\bottomrule
\end{tabular}
\label{tab:results}
\end{center}
\end{table}

\begin{figure*}[htbp]
\centering
\begin{subfigure}[b]{0.32\textwidth}
    \centering
    \includegraphics[width=\textwidth]{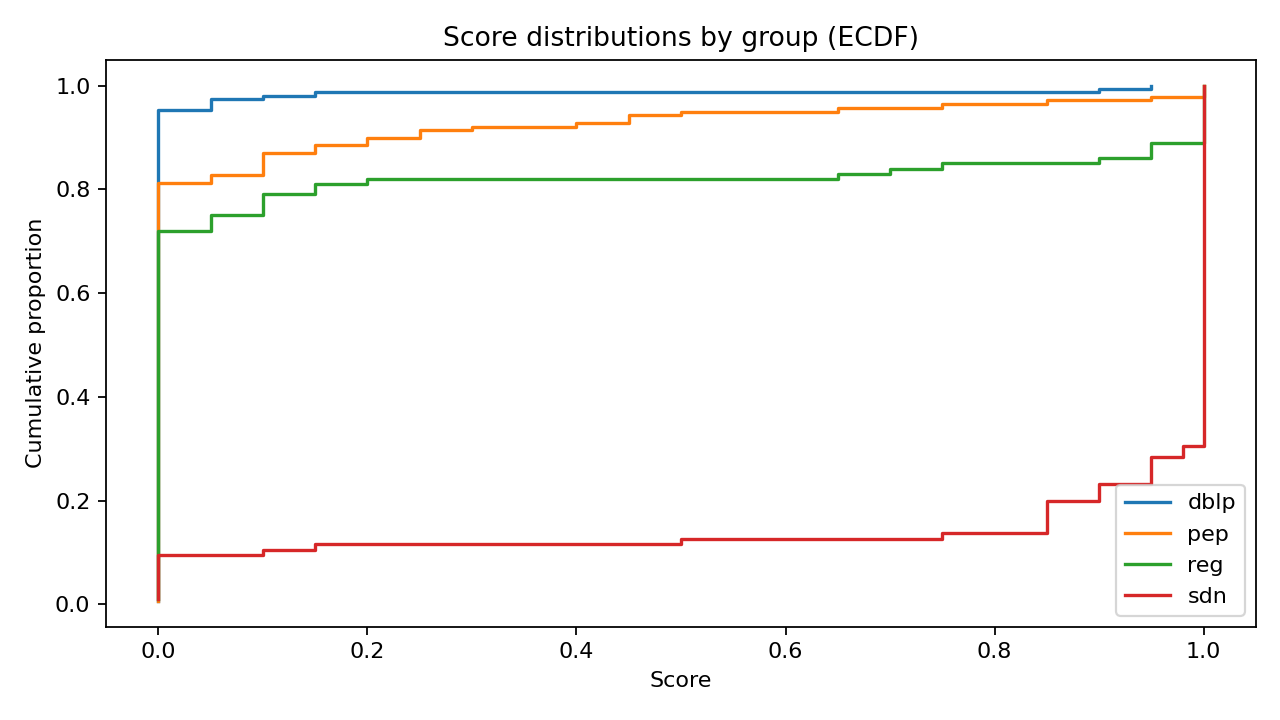}
    \caption{Grok 4.1 Fast}
    \label{fig:ecdf-grok}
\end{subfigure}
\hfill
\begin{subfigure}[b]{0.32\textwidth}
    \centering
    \includegraphics[width=\textwidth]{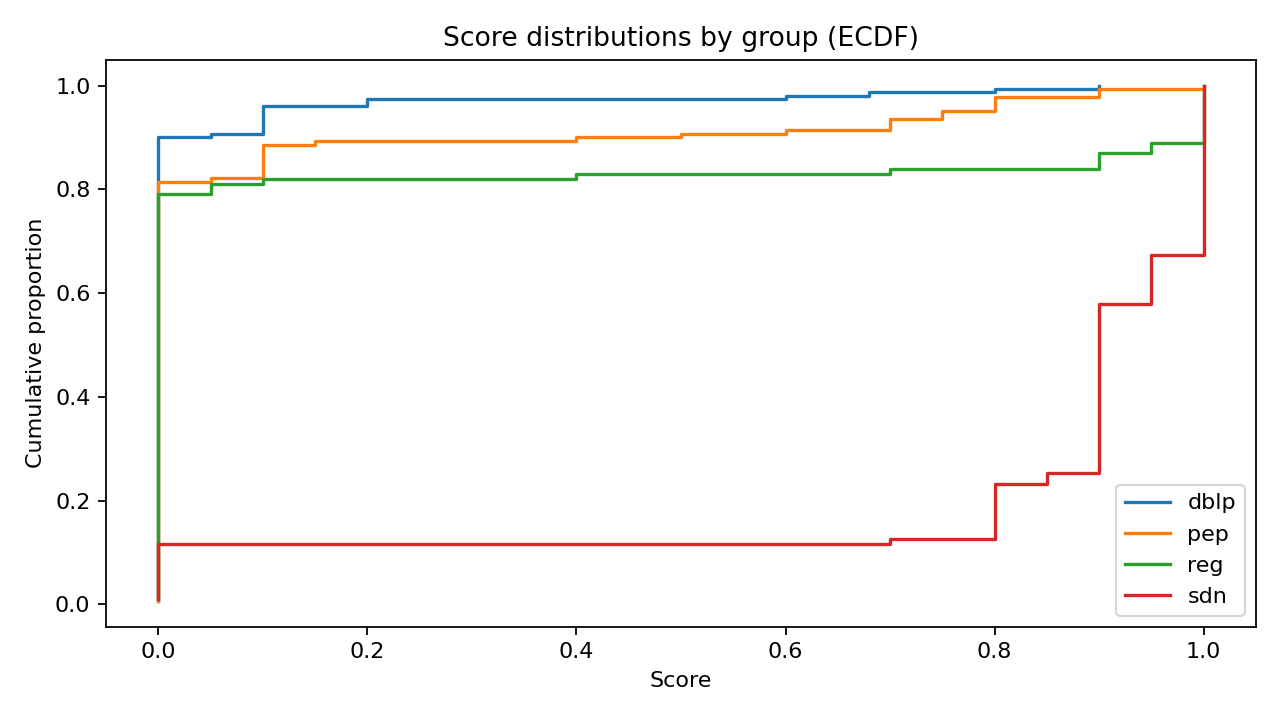}
    \caption{Gemini 2.5 Flash}
    \label{fig:ecdf-gemini}
\end{subfigure}
\hfill
\begin{subfigure}[b]{0.32\textwidth}
    \centering
    \includegraphics[width=\textwidth]{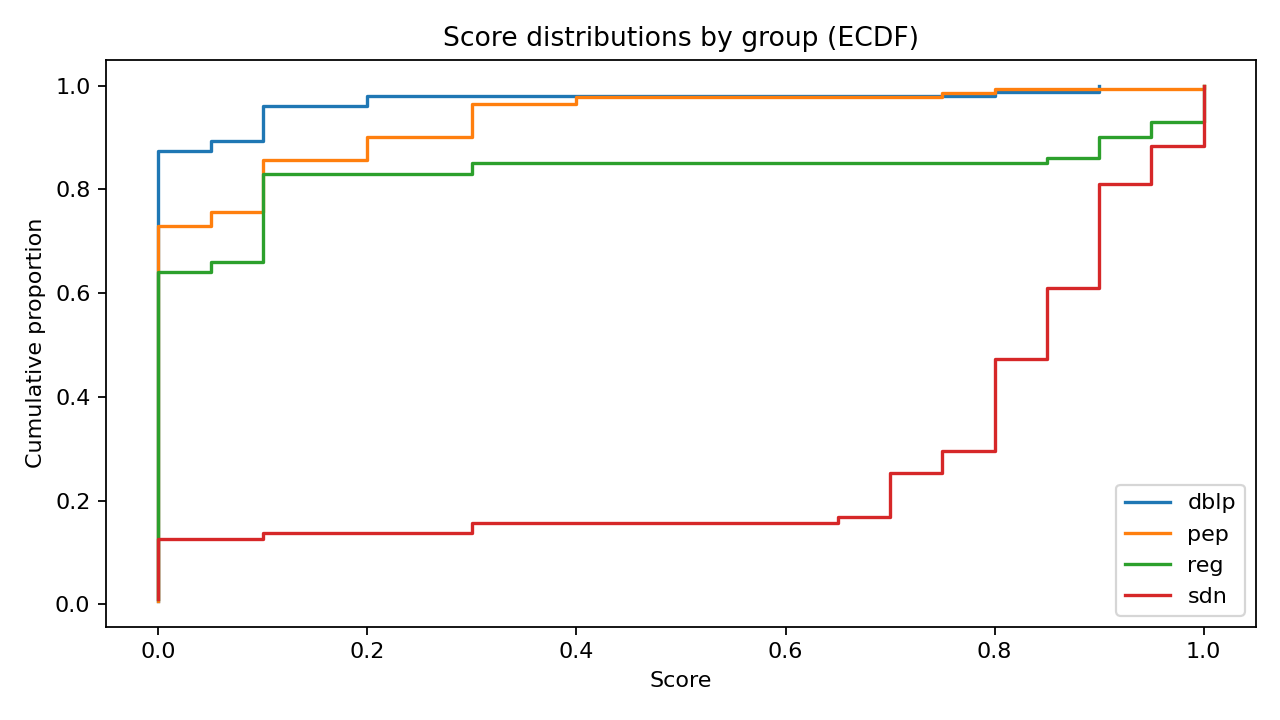}
    \caption{GPT-4.1-Mini}
    \label{fig:ecdf-gpt}
\end{subfigure}
\caption{Empirical Cumulative Distribution Functions (ECDFs) of AMI scores across four populations (Clean, PEP, RW, SDN) for each LLM backend. The clear separation between curves demonstrates the system's ability to discriminate between low-risk and high-risk individuals.}
\label{fig:ecdf}
\end{figure*}

Figure~\ref{fig:ecdf} presents the empirical cumulative distribution functions (ECDFs) of AMI scores for each model. The ECDF plots provide a comprehensive view of score distributions across all four populations. Across all three models, we observe consistent patterns: Clean names cluster tightly near zero, PEP and RW populations show moderate spread in the low-to-mid range, while SDN entries exhibit substantially higher scores with greater variance.

Notably, Grok 4.1 Fast achieves the strongest separation between Clean and SDN populations, with the steepest ECDF curves for low-risk individuals and the highest mean score (0.863) for SDN entries. GPT-4.1-Mini shows more conservative scoring overall, with a lower SDN mean of 0.730, potentially reducing false positives at the cost of some sensitivity.

%% file: sections/06_discussion.tex
\section{Discussion}
\label{sec:discussion}

\subsection{Advantages of the Agentic Approach}

Our LLM-based approach offers several compelling advantages over traditional adverse media screening methods. Perhaps most significantly, large language models bring genuine contextual understanding to the screening process. Unlike keyword-based systems that trigger on surface-level pattern matches, LLMs can interpret nuance and disambiguate meaning based on surrounding context. A mention of ``arrested'' in the phrase ``arrested development of the project'' is correctly distinguished from ``arrested for fraud'', a distinction that would confound simpler approaches and generate false positives.

The interpretability of our system represents another substantial benefit for compliance workflows. Every numeric score produced by the AMI Agent is accompanied by a natural language justification explaining the reasoning behind the assessment. This transparency enables human reviewers to quickly evaluate the system's logic, identify potential errors, and make informed decisions about escalation. For audit and regulatory purposes, this documentation trail provides defensible evidence of due diligence.

The playbook-based architecture provides operational flexibility that traditional systems often lack. Compliance requirements vary across jurisdictions, risk categories, and institutional policies; our configurable playbooks allow practitioners to adapt assessment criteria without retraining models or modifying core system logic. New risk categories can be added, scoring thresholds adjusted, and question formulations refined through simple configuration changes. Furthermore, the system operates in a zero-shot manner, requiring no labeled training data. By leveraging the pre-trained knowledge and reasoning capabilities of foundation models, organizations can deploy effective screening without the substantial upfront investment typically required for machine learning approaches.

\subsection{Limitations and Challenges}

Despite these advantages, several limitations warrant careful consideration. The system's effectiveness fundamentally depends on search engine results, if relevant adverse media is not indexed by search engines or does not rank sufficiently highly in results, it will be missed entirely. Search engine optimization practices, content freshness, and the inherent biases of ranking algorithms all affect coverage in ways that are largely outside our control.

Entity disambiguation remains a persistent challenge, particularly for individuals with common names. While LLMs show promise for zero-shot entity matching by leveraging contextual cues such as associated organizations, locations, and time periods, ambiguous cases inevitably arise. The system may conflate distinct individuals who share names, or conversely fail to connect information about the same person appearing under different name variations. Additional identity attributes such as date of birth or known affiliations can help, but perfect disambiguation remains elusive.

The well-documented tendency of LLMs to hallucinate poses risks in a compliance context. Models may generate plausible-sounding but factually incorrect justifications, potentially leading to erroneous risk assessments. Our RAG architecture mitigates this risk by grounding responses in retrieved documents, but vigilance remains necessary. Human review of justifications, particularly for high-stakes decisions, provides an essential safeguard. Additionally, our current implementation focuses exclusively on English-language sources, limiting coverage for entities whose adverse media appears primarily in other languages. Extending to multilingual screening would require appropriate embedding models and potentially multilingual LLMs, along with careful validation across language pairs.

Finally, temporal considerations affect the relevance of adverse media findings. Historical issues may be less pertinent than recent ones for assessing current risk, yet our system treats all retrieved content equally without weighting by recency. A more sophisticated approach might discount older information or flag the temporal distribution of findings to reviewers.

\subsection{Regulatory Considerations}

The deployment of AI systems for compliance decisions raises important regulatory considerations that practitioners must navigate carefully. Under the EU AI Act, systems used for creditworthiness assessment and similar evaluations may be classified as high-risk, triggering requirements for conformity assessments, human oversight mechanisms, and detailed documentation~\cite{euaiact2024}. While adverse media screening is not explicitly enumerated in the regulation, its role in customer due diligence decisions suggests that cautious interpretation is warranted.

Our system is deliberately designed to augment rather than replace human decision-making. The interpretable justifications facilitate meaningful human review, and the configurable playbooks allow alignment with institutional risk policies and regulatory expectations. However, deployment in regulated financial services contexts requires careful consideration of explainability requirements, fairness implications, and oversight mechanisms. Organizations should conduct thorough impact assessments and maintain clear documentation of how the system's outputs inform, but do not determine, ultimate compliance decisions.

\subsection{Future Directions}

Several promising directions emerge for extending this work. A natural next step involves fine-tuning specialized models on adverse media assessment tasks, which could improve both accuracy and efficiency compared to prompting general-purpose LLMs. Domain-specific training data, potentially generated through our current system with human validation, could enable smaller, faster models that maintain assessment quality while reducing operational costs.

Beyond point-in-time screening, extending the system to continuous monitoring represents a significant opportunity. Rather than conducting periodic batch assessments, a production system could maintain persistent watches on entities of interest, generating alerts when newly published adverse media appears. This would require integration with news aggregation services and efficient incremental processing of new content.

The current architecture could also benefit from multi-source integration, combining unstructured web content with structured data sources such as official sanctions lists, court records, and corporate registries. Such integration would enable cross-validation of findings and provide additional context for risk assessment. Finally, ensuring that numeric AMI scores are well-calibrated probabilities remains an open challenge; principled calibration would enable more defensible threshold selection and facilitate integration with broader risk management frameworks.

%% file: sections/07_conclusion.tex
\section{Conclusion}
\label{sec:conclusion}

We presented an agentic approach to adverse media screening that leverages LLMs with RAG for automated assessment. Our system addresses key limitations of traditional approaches by providing contextual understanding, interpretable justifications, and flexible configuration.

The AMI Agent implements a complete pipeline from web search through verdict generation, producing quantitative risk scores grounded in retrieved evidence. The modular architecture supports multiple LLM backends, enabling deployment across different cost and capability requirements.

As regulatory frameworks evolve from AMLD to AMLR, the need for adaptable, transparent compliance systems will only grow. Our work demonstrates that LLM agents offer a promising path toward more effective and efficient adverse media screening, while maintaining the interpretability essential for regulatory compliance.

The system is available as open-source software, enabling further research and customization for specific compliance requirements.